\documentclass[lettersize,journal]{IEEEtran}
\usepackage{amsmath,amsfonts}
\usepackage{algorithmic}
\usepackage{color}
\usepackage{booktabs}
\usepackage{multirow}
\usepackage{supertabular}
\usepackage{gensymb}
\usepackage{amsmath}
\usepackage{makecell}
\usepackage{leftidx}
\usepackage{flushend}
\usepackage{threeparttable}

\usepackage{array}
\usepackage[caption=false,font=normalsize,labelfont=sf,textfont=sf]{subfig}
\usepackage{textcomp}
\usepackage{stfloats}
\usepackage{url}
\usepackage{doi}
\usepackage{verbatim}
\usepackage{graphicx}
\def\BibTeX{{\rm B\kern-.05em{\sc i\kern-.025em b}\kern-.08em
    T\kern-.1667em\lower.7ex\hbox{E}\kern-.125emX}}
\usepackage{balance}

\begin{document}
\title{RHA-Net: An Encoder-Decoder Network with Residual Blocks and Hybrid Attention Mechanisms for Pavement Crack Segmentation}
\author{Guijie Zhu, Zhun Fan, \IEEEmembership{Senior Member, IEEE}, Jiacheng Liu, Duan Yuan, Peili Ma, Meihua Wang, \\ Weihua Sheng, \IEEEmembership{Senior Member, IEEE}, and Kelvin C. P. Wang
\thanks{Guijie Zhu, Zhun Fan, Jiacheng Liu, Duan Yuan, and Peili Ma are with the College of Engineering, Shantou University, and also with the Key Lab of Digital Signal and Image Processing of Guangdong Province, Shantou, 515063, China (e-mail: zfan@stu.edu.cn).}
\thanks{Meihua Wang with the College of Mathematics and Informatics, South China Agricultural University, Guangzhou, 501642, China (e-mail: wangmeihua@scau.edu.cn).}
\thanks{Weihua Sheng is with the School of Electrical and Computer Engineering, Oklahoma State University, Stillwater, OK 74078 USA.}
\thanks{Kelvin C. P. Wang is with the School of Civil and Environmental Engineering, Oklahoma State University, Stillwater, OK 74078 USA.}}



\maketitle

\begin{abstract}
The acquisition and evaluation of pavement surface data play an essential role in pavement condition evaluation. In this paper, an efficient and effective end-to-end network for automatic pavement crack segmentation, called RHA-Net, is proposed to improve the pavement crack segmentation accuracy. The RHA-Net is built by integrating residual blocks (ResBlocks) and hybrid attention blocks into the encoder-decoder architecture. The ResBlocks are used to improve the ability of RHA-Net to extract high-level abstract features. The hybrid attention blocks are designed to fuse both low-level features and high-level features to help the model focus on correct channels and areas of cracks, thereby improving the feature presentation ability of RHA-Net. An image data set containing 789 pavement crack images collected by a self-designed mobile robot is constructed and used for training and evaluating the proposed model. Compared with other state-of-the-art networks, the proposed model achieves better performance and the functionalities of adding residual blocks and hybrid attention mechanisms are validated in a comprehensive ablation study. Additionally, a light-weighted version of the model generated by introducing depthwise separable convolution achieves better a performance and a much faster processing speed with 1/30 of the number of U-Net parameters. The developed system can segment pavement crack in real-time on an embedded device Jetson TX2 (25 FPS). The video taken in real-time experiments is released at https://youtu.be/3XIogk0fiG4. 
\end{abstract}


\section{Introduction}
\label{Intro}
\IEEEPARstart{C}{racking} is the most critical type of distresses in civil engineering structures, existing in a large variety of infrastructures such as bridges, roads, tunnels, dams and others. Many technologies have been proposed for automating pavement surface distress evaluation \cite{wang2000designs}. Recently, vision-based automatic crack detection methods have attracted a lot of attention from both the academy and the industry due to their advantages of being safer, lower costing, more efficient, and more objective \cite{zou2012cracktree, amhaz2016automatic, zhang2017automated, zhang2017efficient, chen2017nb, yang2019feature}.

In the past few decades, researchers have proposed numerous image processing-based methods for automatic crack detection \cite{oliveira2012automatic, yamaguchi2010fast}. Edge detection-based methods \cite{zhao2010improvement} and threshold-based methods \cite{oliveira2009automatic} are among the first group of methods. However, their robustness and performances are highly susceptible to noises and shadows, partly due to relative simple features used in them.

To improve the robustness and performance of the crack detection methods, more complicated features are utilized, such as Local Binary Pattern \cite{quintana2015simplified} and Histogram of Oriented Gradient (HOG) \cite{kapela2015asphalt}. In addition, machine learning-based classifiers are employed to classify these features \cite{shi2016automatic}.

In recent years, deep learning methods have made great strides in the field of computer vision. Zhang \textit{et al.} \cite{zhang2016road} and Cha \textit{et al.} \cite{cha2017deep} propose CNN-based methods to classify each sub-region of the input image as crack or non-crack by using a sliding window. However, the detection results are sensitive to the choice of the fixed window size. To address this problem, RCNN-based \cite{deng2020concrete} and YOLO-based \cite{du2021pavement} models are proposed to generate flexible proposal regions. However, they can only provide approximate locations of cracks. When fine details of cracks are needed, we need to segment cracks at pixel level. 

As one of the most important tasks in computer vision, object segmentation has been widely studied \cite{long2015fully, chen2017deeplab, badrinarayanan2017segnet}. With the development of object segmentation, many segmentation-based methods have been developed for crack detection \cite{zhang2018deep, choi2019sddnet}. Moreover, Yang \textit{et al.} \cite{yang2019feature} and Zou \textit{et al.} \cite{zou2018deepcrack} use feature fusion method to concatenate the hierarchical features, which can improve the performance of networks because low-level features contain more detailed information of cracks and high-level features have more semantic information.

However, to capture more detailed information of tiny cracks, the network need to pay more attention to low-level features. But by doing that, more background interference will be inevitably introduced because the intervening information (e.g. noises etc) also exists in low-level features. Hence, it is difficult to achieve an excellent trade-off between improving the representation for tiny crack features and suppressing the background interference. In order to solve the aforementioned problems, we propose an improved U-shaped encoder-decoder network for pavement crack segmentation. Specifically, the residual blocks and hybrid attention mechanisms are designed and integrated into U-shaped architecture to improve the performance of the model. Furthermore, the depthwise separable convolution is utilized to reduce the parameters of the model so as to boost its computing efficiency, therefore it can be deployed in real-time.
 
The main contributions of this work are summarized as follows:

\begin{enumerate}
\item We built up a crack image dataset including 789 crack images obtained by a self-designed mobile robot from the campus of Shantou University. The crack images in this dataset contain various backgrounds, such as water stains, oil stains, leaves, branches, pavement markings, shadows, occlusion, soil, sand, and other debris.
\item A novel U-shaped encoder-decoder neural network, namely RHA-Net for pavement crack segmentation, is proposed by integrating residual blocks and hybrid attention mechanisms into an encoder-decoder architecture. The experimental results show that the proposed model achieves better performance with higher computing efficiency than state-of-the-art methods.
\item In order to facilitate practical deployment of the model, we use depthwise separable convolution to largely reduce the parameters and inference time of the model. As a result, the trained light-weighted version of the model can be transplanted into the embedded devices to achieve real-time crack segmentation.
\end{enumerate}

The remainder of this paper is organized as follows. A brief overview of the related works on crack segmentation is introduced in Section \ref{sec:DL}. Section \ref{sec:Method} describes the overall architecture and configurations of the proposed model. Section \ref{sec:Experiments} gives the details of the experiments and discusses the results. We conclude this work in Section \ref{sec:Conclusion}.

\section{{Related} Work}
\label{sec:DL}
In this section, we first briefly review three categories of crack segmentation methods, and then introduce the attention mechanism, which is useful for crack segmentation.

\subsection{Traditional methods}

Based on the characteristic that a crack is darker than its surrounding pixels, it is intuitive to leverage threshold-based methods to segment cracks from their background \cite{oliveira2009automatic, kamaliardakani2016sealed, yamaguchi2008image}. However, threshold-based methods are sensitive to shadows, noises and uneven illuminations, often resulting in discontinuous crack segmentation results. In order to suppress noises and enhance the continuity of the detected cracks, Minimal Path Selection (MPS) is proposed for crack detection \cite{amhaz2016automatic, amhaz2014new}. But still, the performances of MPS-based methods are not guranteed to be satisfactory, especially when the topology of cracks is complex.
Since cracks are similar to edges, many methods based on edge detection have been adopted for crack detection, such as Canny \cite{zhao2010improvement}, Sobel \cite{ayenu2008evaluating} and matched filtering algorithm \cite{zhang2013matched}. Nevertheless, the edges are easily disturbed by background noises, and the hyperparameters of these methods are difficult to be adjusted for different image sets. As a result, these methods based on traditional image processing have difficulties to be applied for solving practical engineering problems.

\subsection{Machine learning-based methods}

With the development of machine learning, methods focusing on feature extraction and pattern recognition have been proposed for crack detection \cite{oliveira2012automatic, cord2012automatic}. Cord \textit{et al.} \cite{cord2012automatic} used AdaBoost to select textural descriptors that can describe crack images. Shi \textit{et al.} \cite{shi2016automatic} proposed a new descriptor based on random structured forests to characterize cracks. The performances of these methods are competitive but very dependent on the extracted features. However, due to complex pavement conditions, it is hard to find common features effective for all pavements. To alleviate this problem, researchers also exploited the features based on the texture of local regions. Michael \textit{et al.} \cite{o2013texture} calculated the feature vector based on the pixel intensity values. Kapela \textit{et al.} \cite{kapela2015asphalt} extracted the HOG features, and LBP features were extracted in \cite{quintana2015simplified, gavilan2011adaptive}. Examples of classifiers include artificial neural network \cite{zakeri2013multi}, and support vector machine \cite{quintana2015simplified, o2013texture}.

\subsection{Deep learning-based methods}
In recent years, convolutional neural networks (CNNs) have achieved unprecedented successes in the field of computer vision \cite{li2021survey}, which include many successful applications of CNNs in crack segmentation tasks. Fan \textit{et al.} \cite{fan2018automatic} proposed a CNN-based method to learn the structure of the cracks from crack images. In another work, Fan \textit{et al.} \cite{fan2020ensemble} constructed an ensemble CNN that consists of three VGG-nets, which votes for a crack segmentation result. Zhang \textit{et al.} \cite{zhang2017automated} introduced a CNN architecture called CrackNet for pixel-level crack segmentation. To prevent information loss, CrackNet contains no pooling layer, and the size of the feature maps is invariant through all layers. Afterwards, Zhang \textit{et al.} developed CrackNet II \cite{zhang2018deep} and CrackNet V \cite{fei2019pixel} based on CrackNet, and achieved higher accuracy and speed. Choi \textit{et al.} \cite{choi2019sddnet} proposed a CNN for segmenting concrete cracks in images. They introduced separable convolution and dilated convolution to design a lightweight crack segmentation network, resulting in a significant reduction in detection time.

In 2015, Long \textit{et al.} \cite{long2015fully} developed a fully convolutional network (FCN) for semantic image segmentation. FCN replaced the fully connected layers at the end of the traditional CNN with convolutional layers, so that the network had no limit on the input size, which greatly promoted the development of segmentation tasks. Yang \textit{et al.} \cite{yang2018automatic} applied FCN to predict segmentation results of cracks. 

The encoder-decoder architecture was one of the most common backbone applied in image segmentation models, like as U-Net \cite{ronneberger2015u} and its variants. Cheng \textit{et al.} \cite{cheng2018pixel} and Jenkins \textit{et al.} \cite{jenkins2018deep} used U-Net for pixel-level crack segmentation. In the upsampling procedure of U-Net, the skip connections between encoder and decoder can help restore information loss to achieve pixel-wise image segmentation \cite{ronneberger2015u}. Ju \textit{et al.} \cite{huyan2020cracku} proposed CrackU-Net, which increases the depth of U-Net and outperforms the original U-Net in pavement crack detection. Liu \textit{et al.} \cite{liu2020automated} applied a modifed U-Net for pixel-level crack segmentation. Bang \textit{et al.} \cite{bang2019encoder} introduced a deep convolutional encoder-decoder network for pixel-level road crack segmentation in images captured by a black-box camera, whose architecture is similar to U-Net. Zou \textit{et al.} \cite{zou2018deepcrack} built DeepCrack net that fuses multi-scale convolutional features from hierarchical convolutional stages for crack segmentation. Fan \textit{et al.} \cite{fan2020automatic} proposed a U-hierarchical dilated network (U-HDN) for crack segmentation, in which the dilated convolution was integrated into the hierarchical feature learning to improve model performance. Yang \textit{et al.} \cite{yang2019feature} proposed a network named FPHBN that utilizes feature pyramid and hierarchical boosting network to incorporate context information from top to bottom. DeepCrack, U-HDN and FPHBN are unlike those methods that only use the last layer of feature maps to classify the pixels. Their feature maps at each scale are integrated to make classifications, preventing information loss signifcantly.

\begin{figure*}[!t]
\centering
\includegraphics[width=18.3cm]{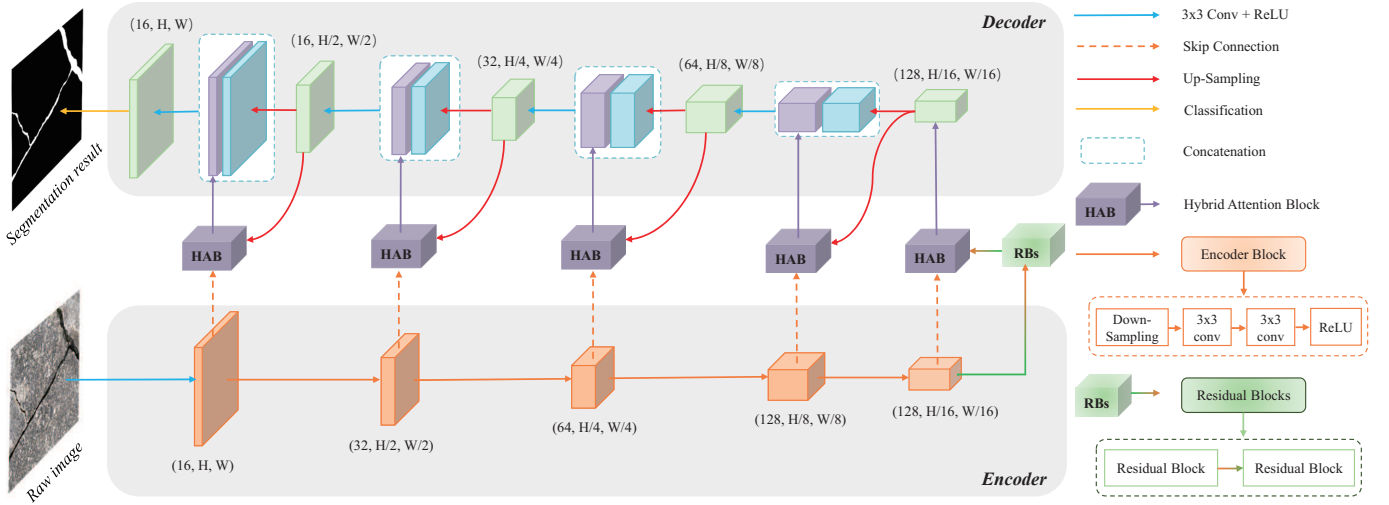}
\caption{Architecture of the proposed RHA-Net.}
\label{fig:RHA}
\end{figure*}

\subsection{Attention mechanism}
In recent years, attention mechanism has gained great interests in deep learning community \cite{9686612, Hu_2018_CVPR, dai2021dynamic, oktay2018attention}. In \cite{Hu_2018_CVPR}, Hu \textit{et al.} proposed squeeze-and-excitation network (SENet), which introduced the attention mechanism into the channel of the feature map. Specifically, the network designed a squeeze-and-excitation operation to make full use of the information between feature channels. Woo \textit{et al.} \cite{woo2018cbam} proposed a convolutional block attention module (CBAM) which contains a channel attention module and a spatial attention module. However, the attention mechanism proposed in the above works is only used to process a single feature map of the input. Oktay \textit{et al.} \cite{oktay2018attention} proposed Attention U-Net model, which can effectively fuse information from both low-level feature maps and high-level ones. They further embedded the attention gate blocks into the U-Net architecture to improve feature representation in regions of interest. In the crack segmentation task, Qu \textit{et al.} \cite{9525316} integrated the Res2Net backbone with the attention mechanism and multi-features fusion for pavement crack segmentation. Liu \textit{et al.} \cite{9686612} proposed a feature fusion module based on attention mechanisim to improve the representational capacity of the proposed model for crack features. Sun \textit{et al.} \cite{9741463} adopted the DeepLabv3+ model and enhanced it for crack segmentation, in which a multi-scale attention module was embeded into the encoder of DeepLabv3+ to help the model obtain more accurate crack segmentation results. In another work, Qu \textit{et al.} \cite{9712196} utilized DCANet backbone network and designed an attention module by integrating the low-level features and the high-level features to recover the crack edge information. In short, the attention mechanism can facilitate fusing feature information from multiple layers. Inspired by them, we designed a novel hybrid attention block by merging the channel-wise attention and the spatial-wise attention to better fuse information from low-level and high-level feature maps.

\section{Proposed Method}
\label{sec:Method}
In this section, we explain the proposed method in more details, including its overall architecture, the two important building blocks, the light-weighted version, and the loss function used in this work.

\subsection{Network Architecture}
RHA-Net has an encoder network, two residual blocks, a corresponding decoder network, five hybrid attention blocks, and a pixelwise classification layer. 
The architecture of RHA-Net is shown in Fig. \ref{fig:RHA}. The encoder network consists of an initial convolution block and four encoder blocks. 
The original image is processed by a 3$\times$3 convolution operation and a rectified linear unit (ReLU), and then pass through four encode blocks. 
The encoder block is a series of down-sampling operation of maxpooling and two repeated applications of 3$\times$3 convolution which reduce the image size to half of the original size and double the channels of the feature maps. 
In the deepest layer of the RHA-Net, we use the residual blocks to deepen the network, i.e. to increase the number of network layers and avoid network degradation. 
The decoder network is designed according to the encoder network. Each decoder block consists of a combination of corresponding operations of a bilinear up-sampling and a 3$\times$3 convolution operation.
Combining the output results of decoder and the corresponding encoder effectively enhances the ability of the network to segment image details. 
To achieve a better combination, we designed a hybrid attention block (HAB) integrating two types of attentions, including channel attention and spatial attention. 
The HABs are embedded into the encoder-decoder architecture so that the outputs from corresponding pairs of encoder and decoder are combined in an optimized way.  
The output of the decoder is fed to a softmax classifier to predict the segmentation result.

\subsection{Residual Block}
\begin{figure}[t]
	\centering
	\includegraphics[width=7cm]{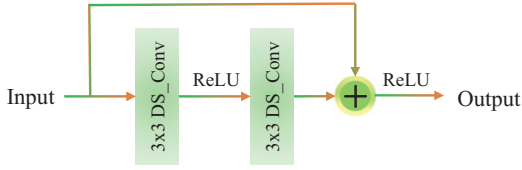}
	\caption{Residual block with depthwise separable convolution.}
	\label{fig:resb}
\end{figure}

Residual block is first proposed and applied to ResNet \cite{he2016deep} to solve the degradation problem of deep convolution network. 
When the network reaches a certain depth, the performance of the model may fall into a local optima and is difficult to improve. 
When the network continues to deepen, the performance of the model on the test set may even decrease. This phenomenon is called network degradation. 
Because residual blocks can help to alleviate the problem of network degradation, we use two repeated applications of residual blocks. 
To make the model more light-weighted, we replace the conventional convolution with depthwise separable convolution (DS\_Conv) and integrate this light-weighted residual blocks into the proposed RHA-Net.
The configuration of light-weighted residual blocks is shown in Fig.~\ref{fig:resb}. 

\begin{figure*}[!t]
\centering
\includegraphics[width=18.2cm]{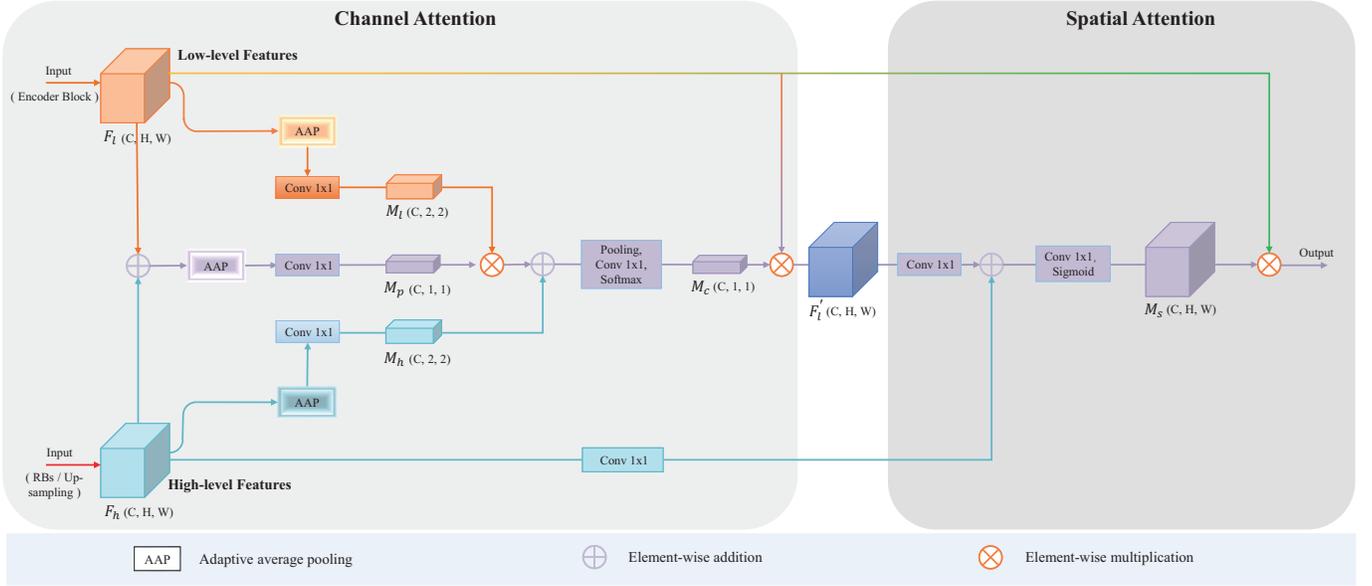}
\caption{Schematic of the proposed hybrid attention block.}
\label{fig:HA}
\end{figure*}

\subsection{Hybrid Attention Block}
Since the Encoder-Decoder architecture produces hierarchical features in the encoder and the decoder, 
we design the hybrid attention block using the low-level features $ F_{l} $ (the output of Encoder blocks) and the high-level features $ F_{h} $ (the output of residual blocks or up-sampling).

Generally speaking, the information presented in the low-level features focuses more on the location details of the cracks, while the high-level features contain more semantic information of the cracks. 
Inspired by CBAM \cite{woo2018cbam}, we choose to fuse low-level features and high-level features using both channel and spatial attentions.
Channel attention tells us which channels' outputs should be more emphasized, while spatial attention specifies which positions' outputs should be more valued.  
Specifically, the channel attention focuses on 'what' is meaningful for the input image and the spatial attention focuses on 'where' is meaningful for the input image. 
A hybrid attention block uses low-level features and its corresponding high-level features as input. 
The configuration of hybrid attention block is shown in Fig.~\ref{fig:HA}.

On the channel part, inspired by SENet \cite{Hu_2018_CVPR}, adaptive average pooling, full connection layer and activation function are used to calculate the channel attention map which enables the network to distinguish the importance of different channels and improve the expression ability of the network.  
The low-level features and the corresponding high-level features have the same size, we can therefore add the two feature maps pixel by pixel to get an overlapping map, then use adaptive average pooling operation to squeeze this map into a map of the channel weight and use  1$\times$1 convolution to extract the relationship between channels. 
This operation turns every two-dimensional feature map of each channel into a weight number with global receptive field, which means much channel information is lost and this map of the channel weights only has preliminary channel attention information. 
Because the low-level features and the high-level features have different image performance capabilities, the preliminary channel attention map $ M_{p} $ needs to fuse these two features to supplement its channel attention information. 
Compared with the high-level features, the low-level features generated by encoder pass through fewer convolution operations, which means the low level feature map $ M_{l} $ squeezed by adaptive average pooling and 1$\times$1 convolution operation has less semantic information. 
Correspondingly, the high-level feature map $ M_{h} $ has more semantic information. 
Therefore, we multiply $ M_{p} $ with  $ M_{l} $ to downplay unimportant channels and highlight the important ones, then fuse the result with $ M_{h} $ by element-wise addition. 
Through the above operations, channel attentions of low-level features and high-level features are more fully utilized, and channel attention map $ M_{c} $ is obtained after average pooling, 1$\times$1 convolution and softmax operations in sequence.
In short, the channel attention map is computed as follows:

\begin{equation}
  \begin{split}
    M_{c} =& \mathit{f_{c}}(F_{l}, F_{h}) = \mathit{Softmax}(\sigma (\mathit{Avg Pool} (\sigma(AAP(\\
    &F_{h} \oplus F_{l})) \otimes \sigma(\mathit{AAP}(F_{l}))\oplus \sigma(\mathit{AAP}(F_{h})))))\\
    =& \mathit{Softmax} ( \mathit{AvgPool}( M_{p} \otimes  M_{l} \oplus M_{h} ) )
  \end{split}
  \end{equation}
where $ \mathit{f_{c}}(F_{l}, F_{h} ) $ is the calculation function of channel attention map, $ \sigma $ represents a 1$\times$1 convolution operation and a rectified linear unit, $ \mathit{AAP} $ represents an adaptive average pooling operation, $\mathit{Avg Pool}$ represents an average pooling operation, $\mathit{Softmax}$ represents a softmax operation, $\oplus$ denotes the element-wise addition, and $\otimes$ denotes the element-wise multiplication.
The final output of channel part $ F_{l}^{'} $ is obtained by multiplying the channel attention map $ M_{c} $ with the low-feature $ F_{l} $. 

On the spatial part, extracting region of interests from the feature map facilitates filtering out the influence of the background and improve the spatial expressive ability of the network. 
We use element-wise addition to fuse the $ F_{l}^{'} $ and $ F_{h} $, then use 1$\times$1 convolution operation and sigmoid function to extract the interested spatial area and generate the corresponding spatial attention map.
In short, the spatial attention is computed as follows:
\begin{equation}
  \begin{split}
    M_{s} = \mathit{f_{s}}(F_{l}^{'}, F_{h} ) = \mathit{Sigmoid} (\sigma ( F_{l}^{'}) \oplus \sigma (F_{h}) )
  \end{split}
  \end{equation}
where $ \mathit{f_{s}}(F_{l}, F_{h} ) $ is the calculation function of spatial attention map, and $\mathit{Sigmoid}$ represents an sigmoid function.
Finally, we multiply the spatial attention map with the low-level feature map to get the output features of the hybrid attention module which hybrids channel attention and spatial attention.
In summary, the hybrid attention module is computed as follows:
\begin{equation}
  \begin{split}
    \mathit{f_{hybrid}}(F_{l}, F_{h} ) = \mathit{f_{s}}(F_{l} \otimes \mathit{f_{c}}(F_{l}, F_{h} ), F_{h} ) \otimes F_{l} 
  \end{split}
  \end{equation}

\subsection{The Light-weighted Version of the Proposed Model}

\begin{figure}[!t]
\centering
\includegraphics[width=8cm]{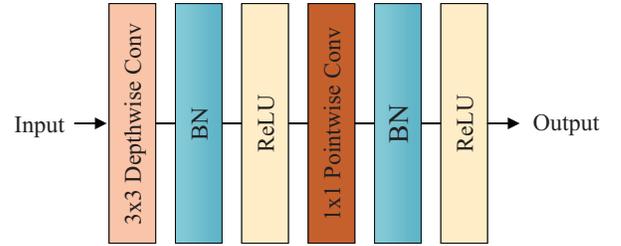}
\caption{Depthwise separable convolution operation.}
\label{fig:DS_Conv}
\end{figure}

To facilitate the deployment of the proposed network to the embedded platform, we replace all the convolution operations in the network with depthwise separable convolution operations, 
which can improve the running speed of the network and reduce its parameters. 
Compared with conventional convolution, the depthwise separable convolution decompose a complete convolution operation into two steps: depthwise convolutions layer (DW) with batch normalization (BN) and rectified linear unit (ReLU), 
pointwise convolutions layer (PW) with BN and ReLU, as shown in Fig.~\ref{fig:DS_Conv}. 
In conventional convolution, each convolution kernel has to convolve all channels, while depthwise convolution only allows each convolution kernel to convolve one channel, 
which means that the amount of computation and parameters of convolution operations are greatly reduced. 
However, if only depthwise convolution is used, the information between channels is not utilized. 
The PW convolution fuse the feature maps of all channels with a 1$\times$1 convolution kernel.

\subsection{Loss Function}
For the crack segmentation task, the main purpose is to distinguish the crack pixels from the background. In a crack image dataset, the number of crack pixels is usually much smaller than that of non-crack ones, which means that the dataset is significantly imbalanced. To adress this problem, we employ the weighted binary cross-entropy as the loss function in the training phase, which is formulated as follows:
\begin{equation}
\begin{split}
						L_{\omega bce} = -\mathop \sum \limits_{i = 1}^{{N}}{(\omega_p  y_i \log(\hat{y}_i)+(1 - y_i)\log(1-\hat{y}_i))}     
\end{split}
\end{equation}
where $N$, $y$, $\hat{y}$, and $\omega_p$ respectively denote the total number of samples, model prediction, ground truth, and balance factor between crack and non-crack samples. 

\section{Experiments And Results}
\label{sec:Experiments}
In this section, we first introduce the self-built crack dataset (CamCrack789), and three other public datasets (Crack500, CFD, and DeepCrack237). Then experimental settings and evaluation metrics are explained. At last, the experimental results are provided and analyzed.  

\subsection{Datasets}
\subsubsection{CamCrack789}
\begin{figure*}[!t]
    \centering
        \includegraphics[width=16cm]{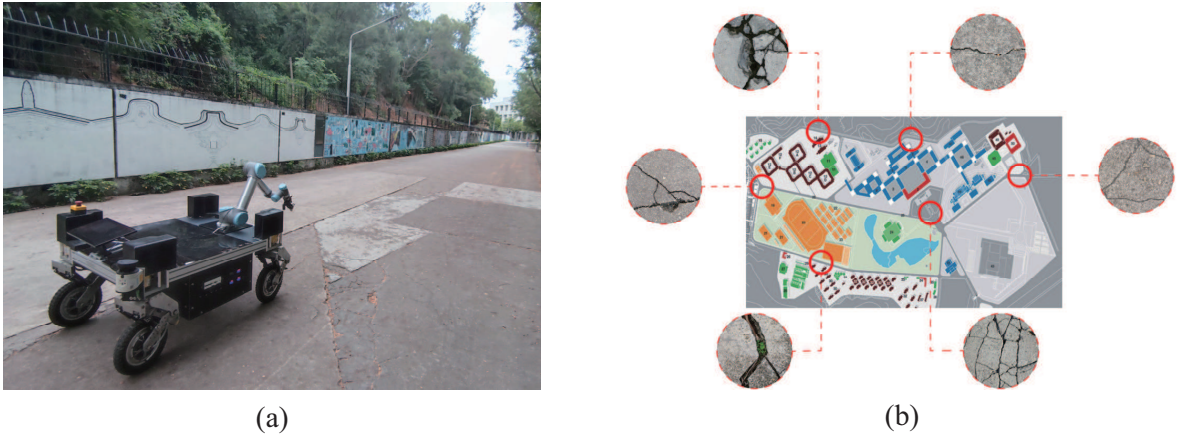}
    \caption{Robot and locations of pavement crack images. (a) Robot for road crack image acquisition and detection. (b) Images taken from different locations.}
    \label{fig:robot}
\end{figure*}

\begin{figure*}[!t]
    \centering
        \includegraphics[width=18cm]{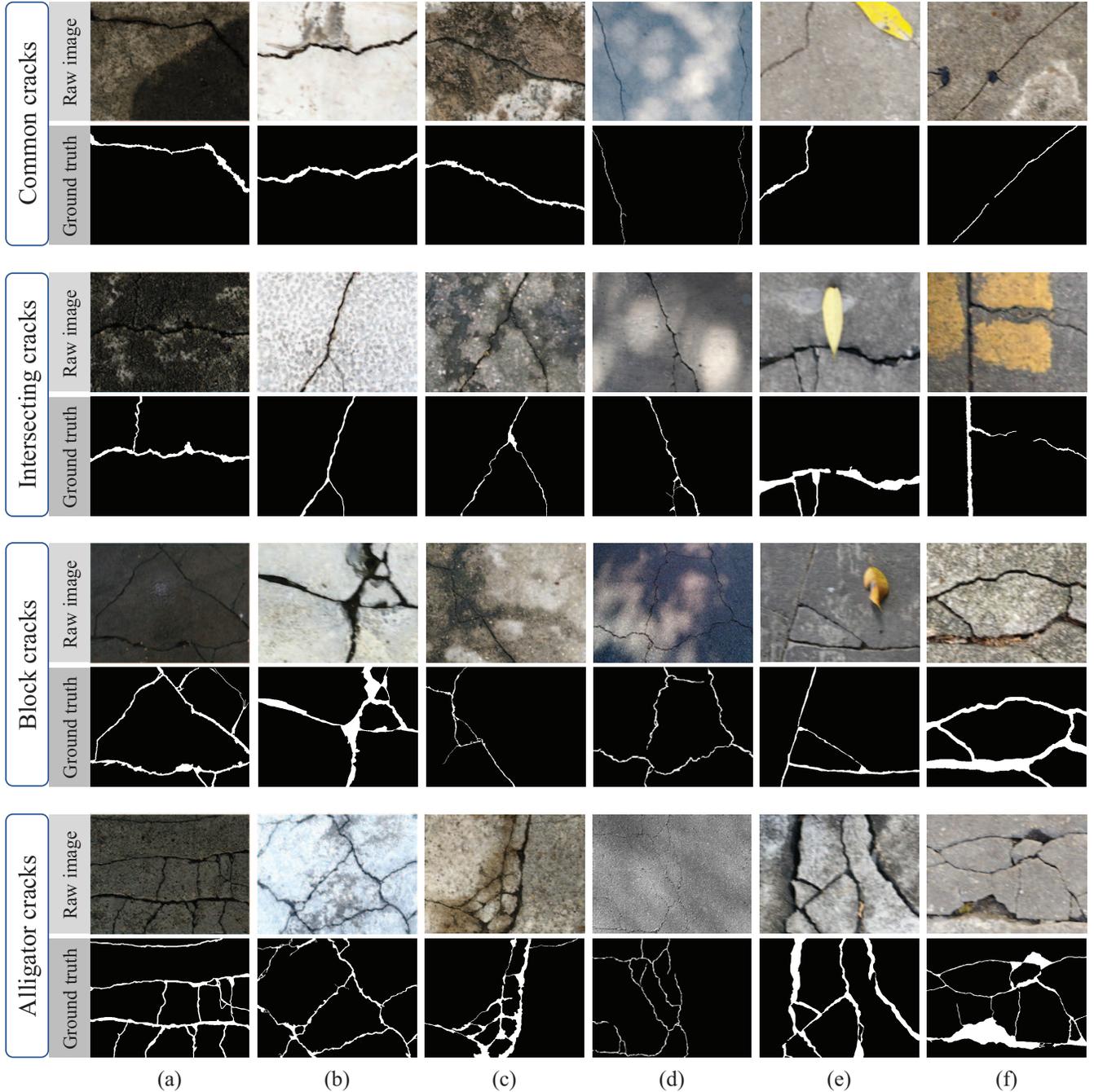}
    \caption{Examples of the CamCrack789 dataset. According to the topology of the cracks, they are divided into four types: common cracks, intersecting cracks, block cracks, and alligator cracks. The crack images are collected under different lighting conditions, such as (a) dark light and (b) bright light. In addition, they contain various backgrounds, such as (c) water stains, (d) shadows, (e) tree leaves, and (f) other backgrounds, including pavement marking, oil stain, debris, etc.}
    \label{fig:example}
\end{figure*}

As shown in Fig.~\ref{fig:robot}, the images of pavement cracks are taken from different locations on the campus of Shantou University. The pictures are captured by a Microsoft HD camera mounted on the end of the manipulator of a self-designed mobile robot capable of working in different weather conditions, including sunny, cloudy, and rainy days. The camera can record 2,304$\times$1,728 pixel video at 30 fps. In total, 789 images of pavement cracks are selected to construct a pavement crack dataset named CamCrack789. It consists of crack images with 640$\times$480 pixels in size containing water stains, oil stains, leaves, branches, pavement markings, shadows, occlusion, soil, and other debris. In addition, some blurred crack images caused by the movement of the mobile robot or poor lighting conditions are included. In CamCrack789, the ground truth of each image is manually labeled at the pixel level. some examples of raw images and corresponding ground truth are provided in Fig.~\ref{fig:example}. This dataset is randomly divided into two parts: 546 images for training and 243 images for testing. 

To enhance the generalization of the models and prevent overfitting of the models, we apply data augmentation method on CamCrack789 to increase the number of traning samples. We randomly rotate the crack images by 180°, flip vertically and horizontally, and change the brightness and contrast of the crack images. 

\subsubsection{Crack500 \cite{yang2019feature}}
The Crack500 dataset includes 3368 pavement crack images from 500 raw crack images with around 2000$\times$1500 pixels captured by cell phone. All the raw images are manually labeled at the pixel level. In order to prompt the image processing speed and reduce the consumption of the computation resorces, each image is resized to 256$\times$256 pixels. In Crack500, 1896 images are used as the training samples, 348 images as the validation samples, and 1124 images as the test samples. 
\subsubsection{CFD \cite{shi2016automatic}}
This dataset contains 118 RGB images with a fixed size of 480$\times$320 pixels captured by an iphone5 in Beijing, China. Each image in CFD is also manually annotated to describie ground truth contours of cracks. These images are randomly divided into two parts at a 6:4 ratio (i.e., 72 images for training and the rest 46 images for testing).
\subsubsection{DeepCrack237}
This dataset is the testing set of the DeepCrack dataset \cite{liu2019deepcrack}, which contains 237 RGB color images manually annotated segmentations. The image resolution is 544$\times$384 pixels. We use this dataset to evaluate the generalization ability of models.


\subsection{Experimental Settings}

The proposed method for crack segmentation is implemented using the widely used PyTorch library. The specifications of the computing platform used for training and testing the network were as follows: Intel(R) Xeno(R) Gold 5115 2.40GHz CPU, NVIDIA 16G GeForce RTX 2080 Ti. In the training phase, Adam \cite{Kingma2014Adam} is adopted as an optimizer with a learning rate of 1e-3 to optimize paramters of the networks. When training the models, batch size is set to 8 for CamCrack789 and Crack500, and 12 for CFD, the number of training epoch is set to 500. The models are saved every 50 epochs. 

\subsection{Evaluation Metrics}
Pavement crack segmentation is a binary classification problem at pixel-level. We need to classify each pixel in  the crack image as crack or non-crack. The prediction output of the model is a classification probability map, i.e., each pixel in the map is assigned a probability of belonging to the class of cracks.
Note that in all experiments, the results are obtained with the probability threshold set to 0.5. In this work, the following three metrics are adopted for evaluating the model: Precision ($Pr$), Recall ($Re$), and F1 score ($F1$). They are defined as follows:
\begin{equation}
\begin{split}
Pr= \frac{TP}{TP+FP}
\end{split}
\end{equation}
\begin{equation}
\begin{split}
Re= \frac{TP}{TP+FN}
\end{split}
\end{equation}
\begin{equation}
\begin{split}
F1= \frac{2\times Pr \times Re}{Pr+Re}
\end{split}
\end{equation}
where $TP$, $TN$, $FP$, $FN$ are the numbers of true positive, true negative, false positive, and false negative, respectively. For crack detetion research, considering the cracks width, a small distance (2 pixels in \cite{amhaz2016automatic, zhou2022} and 5 pixels in \cite{shi2016automatic}) between the prediction result and the ground truth is allowed in evaluation. We adopt 2 pixels in this study.

\subsection{Experimental Results}


We used a U-shaped encoder-decoder architecture as the backbone in this work and compared the proposed network with five re-implemented methods, which contain FCN, U-Net, Attention U-Net, DeepCrack \cite{zou2018deepcrack}, and DeepCrack \cite{liu2019deepcrack}. For a fair comparison, we trained all these networks with the same hyperparameters. We report the evaluation results of the proposed method on four datasets and compare them with other existing methods.
\subsubsection{Results on CamCrack789}
\begin{figure}[!t]
	\centering
	\includegraphics[width= 8.9cm]{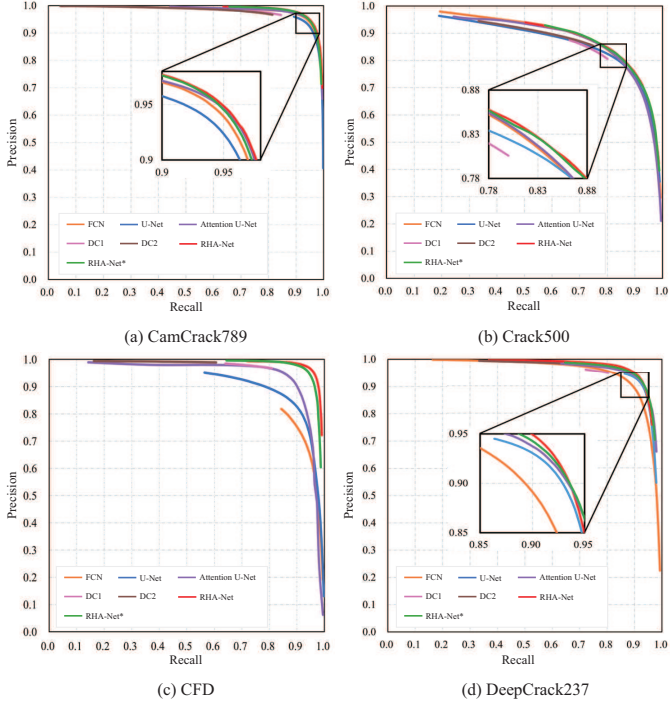}
	\caption{P-R curves of compared methods on the CamCrack789, Crack500, CFD, and DeepCrack237 respectively.} 
	\label{fig:PR}
\end{figure}

\begin{figure*}[!t]
	\centering
	\includegraphics[width=18cm]{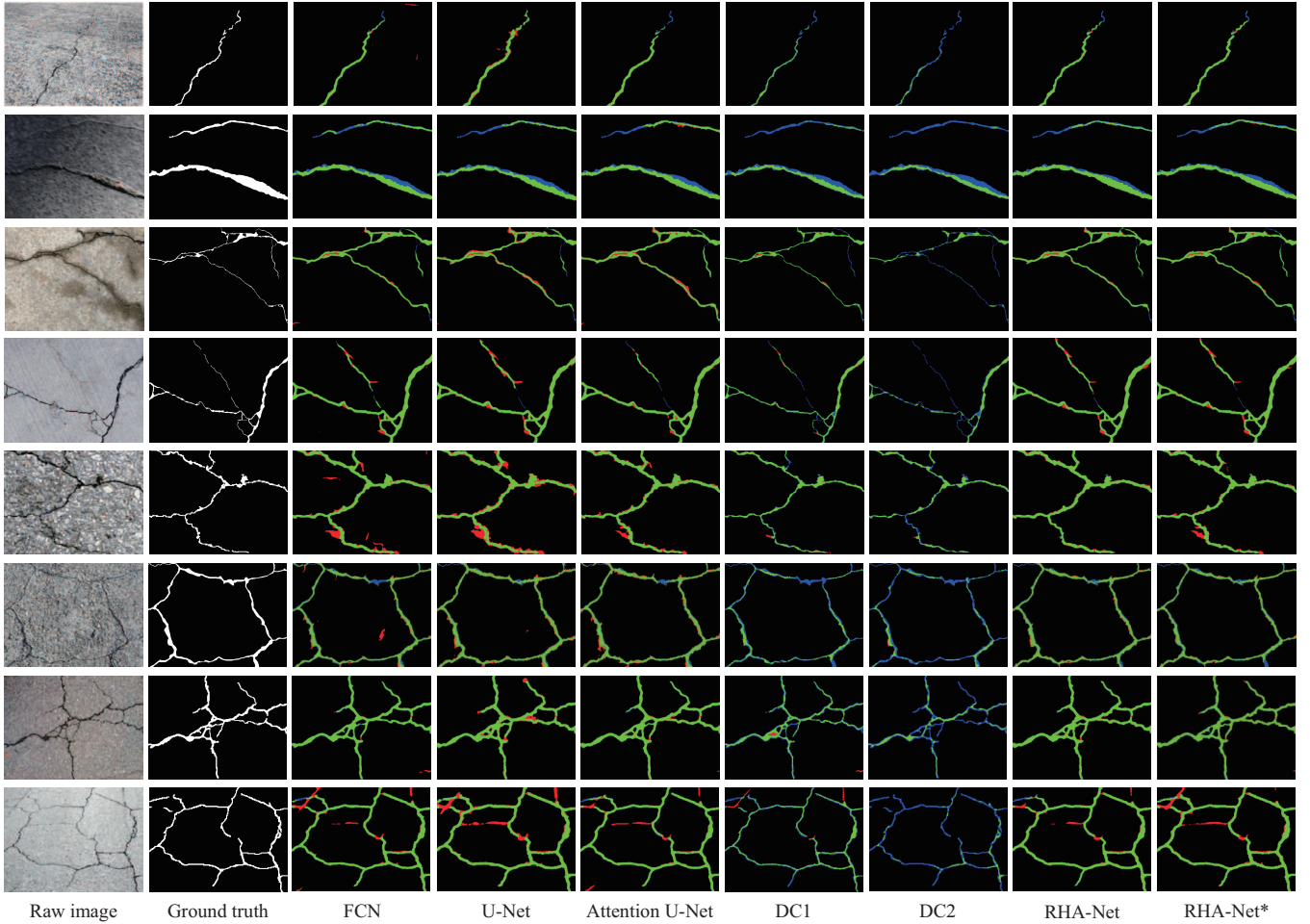}
	\caption{Visualization results of different methods selected from CamCrack789. DC1 and DC2 mean the methods of DeepCrack \cite{zou2018deepcrack} and DeepCrack \cite{liu2019deepcrack} respectively. The green pixels represent true positives, red pixels represent false positives, and blue pixels represent false negatives.} 
	\label{fig:result}
\end{figure*}

It can be seen from Table I that the proposed model RHA-Net gets the best results with the F1 95.30\%. 
Compared to FCN, U-Net, Attention U-Net, DeepCrack \cite{zou2018deepcrack}, and DeepCrack \cite{liu2019deepcrack}, the performance improvement on F1 made in RHA-Net are 4.56\%, 4.92\%, 2.53\%, 8.12\% and 22.87\% respectively. Note that the light-weighted version of RHA-Net is only 0.29\% lower than RHA-Net on F1, but has better performance than other comparison methods.
From Fig.~\ref{fig:PR}(a), we can see that RHA-Net has the best precision and recall values. Furthermore, in visualization results Fig.~\ref{fig:result}, RHA-Net can accurately identify the cracks and suppress the noise of the background, and DeepCrack \cite{liu2019deepcrack} has the worst segmentation results, which demonstrates less continuity than all other methods.

\subsubsection{Results on Crack500}
It can be seen from Table II that RHA-Net outperforms other compared methods on the Crack500 dataset. Compared to FCN, U-Net, Attention U-Net, DeepCrack \cite{zou2018deepcrack}, and DeepCrack \cite{liu2019deepcrack}, it is 4.77\%, 3.20\%, 2.38\%, 4.87\% and 8.25\% performance improvement on F1, respectively. In Fig.~\ref{fig:PR}(b), the light-weighted model RHA-Net* shows similar perfomance to RHA-Net and better than other compared deep learning models, which can also be seen from Fig.~\ref{fig:result2}(a).

\subsubsection{Results on CFD}
As shown in Fig.~\ref{fig:PR}(c), the proposed RHA-Net achieves superior performance than the others on this dataset, and the light-weighted version of RHA-Net gains similar performance to RHA-Net. From Table III, it can be observed that RHA-Net significantly outperforms other comprison methods in terms of Re and F1. The F1 values of FCN, U-Net, Attention U-Net, DeepCrack \cite{zou2018deepcrack} and DeepCrack \cite{liu2019deepcrack} are 32.72\%, 14.00\%, 8.16\%, 13.07\% and 27.80\% lower than our proposed method performance, respectively. Note that although all methods can not completely segment the cracks with complex topological structure, our method can distinguish and segment most of those tiny cracks, as shown in Fig.~\ref{fig:result2}(b).

\subsubsection{Results on DeepCrack237}
\begin{figure*}[!t]
	\centering
	\includegraphics[width=18.2cm]{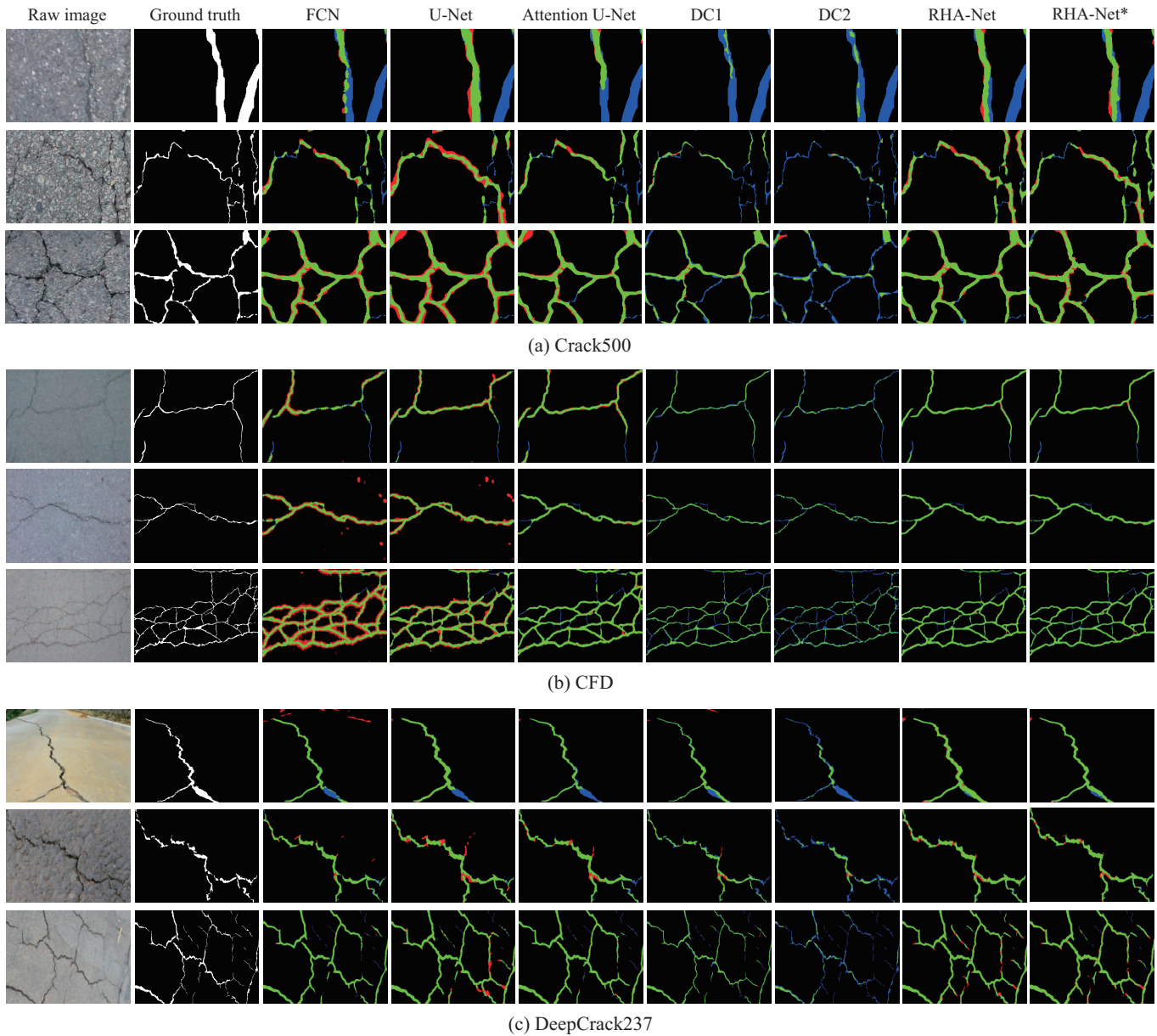}
	\caption{Visualization results of different methods selected from Crack500, CFD, and DeepCrack237, respectively. The green pixels represent true positives, red pixels represent false positives, and blue pixels represent false negatives.} 
	\label{fig:result2}
\end{figure*}

From Fig.~\ref{fig:PR}(d) and Fig.~\ref{fig:result2}(c), we can see that the proposed RHA-Net outperforms other comparison methods on DeepCrack237. From Table IV, we see that RHA-Net and its light-weighted version RHA-Net* outperform Attention U-Net by 2.46\% and 1.87\% in terms of F1, respectively. As all models are trained on CamCrack789 but evaluated by DeepCrack237, the experiment demonstrates that the proposed method has better generalizability than other comparative methods.

\begin{table}
\renewcommand\arraystretch{1.5}
  \centering
  \begin{threeparttable}[b]
  \setlength{\tabcolsep}{2.4mm}{
  \caption{Comparison of segmentation results on CamCrack789  }    
  \begin{tabular}{lcccc}
    \toprule
    Methods  & Pr & Re & F1 & time/image(s) \\
    \midrule
		  FCN \cite{long2015fully} &  0.8757 & 0.9566 & 0.9074 & 0.075 \\
    U-Net \cite{ronneberger2015u} &  0.8586 & {\bfseries 0.9802}  & 0.9038 & 0.111\\ 
    Attention U-Net \cite{oktay2018attention} &  0.8864 & 0.9793 & 0.9277 & 0.199\\
    DeepCrack \cite{zou2018deepcrack}  &  0.9719 & 0.8021 & 0.8718 & 0.361\\
    DeepCrack \cite{liu2019deepcrack}  &  {\bfseries 0.9773} & 0.6011 & 0.7243 & 0.058\\
    \midrule
    RHA-Net  & 0.9445 & 0.9654 & {\bfseries 0.9530} &  0.033 \\
		  RHA-Net$^{*}$ & 0.9416 & 0.9628 & 0.9501 & {\bfseries 0.032} \\
    \bottomrule
  \end{tabular}}
  \label{camcrack789}
   \begin{tablenotes}
    \item[] "*" means the light-weighted version of the model, i.e. the model obtained by using depthwise separable convolution. 
   \end{tablenotes}
  \end{threeparttable}
\end{table}

\begin{table}
\renewcommand\arraystretch{1.5}
  \centering
  \setlength{\tabcolsep}{2.4mm}{
  \caption{Comparison of segmentation results on Crack500}    
  \begin{tabular}{lcccc}
    \toprule
    Methods  & Pr & Re & F1 & time/image(s) \\
    \midrule
			FCN \cite{long2015fully} &  0.6369 & {\bfseries 0.9390} & 0.7416 & 0.017\\
    U-Net \cite{ronneberger2015u} &  0.6683 &  0.9242 & 0.7573 & 0.025\\ 
    Attention U-Net \cite{oktay2018attention} &  0.6753 & 0.9321 & 0.7655 & 0.045\\
    DeepCrack \cite{zou2018deepcrack}  &  0.8286 & 0.7140 & 0.7406 & 0.082\\
    DeepCrack \cite{liu2019deepcrack}  &  {\bfseries 0.8591} & 0.6408 & 0.7068 & 0.015\\
    \midrule
    RHA-Net & 0.7140 & 0.9212 & {\bfseries 0.7893} & {\bfseries 0.011} \\
			RHA-Net$^{*}$ & 0.6919 &  0.9319 &   0.7781 & {\bfseries 0.011} \\
    \bottomrule
  \end{tabular}}
  \label{table:crack500}
\end{table}

\begin{table}
\renewcommand\arraystretch{1.3}
  \centering
  \setlength{\tabcolsep}{2.4mm}{
  \caption{Comparison of segmentation results on CFD}    
  \begin{tabular}{lcccc}
    \toprule
    Methods  & Pr & Re & F1 & time/image(s) \\
    \midrule
			 FCN \cite{long2015fully} &  0.5412 & 0.8055 & 0.6300 & 0.064\\
    U-Net \cite{ronneberger2015u} &  0.7781 & 0.8972  & 0.8170 & 0.068\\ 
    Attention U-Net \cite{oktay2018attention} &  0.8857 & 0.9039 & 0.8756 & 0.117\\
    DeepCrack \cite{zou2018deepcrack}  &  0.9693 & 0.7322 & 0.8265 & 0.200\\
    DeepCrack \cite{liu2019deepcrack}  &  {\bfseries 0.9860} & 0.5392 & 0.6792 & 0.047\\
    \midrule
    RHA-Net & 0.9640& {\bfseries 0.9528} & {\bfseries 0.9572} & {\bfseries 0.030} \\
			 RHA-Net$^{*}$ & 0.9588 & 0.9439 &  0.9496 & {\bfseries 0.030} \\
    \bottomrule
  \end{tabular}}
  \label{table:CFD}
\end{table}

\begin{table}
\renewcommand\arraystretch{1.5}
  \centering
  \setlength{\tabcolsep}{2.4mm}{
  \caption{Results of Compared method test on DeepCrack237}    
  \begin{tabular}{lcccc}
    \toprule
    Methods  & Pr & Re & F1 & time/image(s) \\
    \midrule
	 FCN \cite{long2015fully} &  0.9184 & 0.8863 & 0.8797 & 0.152 \\
    U-Net \cite{ronneberger2015u} &  0.8557 &  0.9442  & 0.8867 & 0.177\\ 
    Attention U-Net \cite{oktay2018attention} &  0.9198 & 0.9086 & 0.9006 & 0.317\\
    DeepCrack \cite{zou2018deepcrack}  &  {\bfseries 0.9516} & 0.7846 & 0.8492 & 0.585\\
    DeepCrack \cite{liu2019deepcrack}  &   0.9482 & 0.5912 & 0.7046 & 0.107\\
    \midrule
    RHA-Net  & 0.9162 & {\bfseries 0.9484} & {\bfseries 0.9252} &  0.083\\
	  RHA-Net$^{*}$ & 0.9247 & 0.9348 & 0.9193 & {\bfseries 0.080} \\
    \bottomrule
  \end{tabular}}
  \label{table:crack500}
\end{table}

\subsection{Ablation Study}
\label{sec: Ablation}
In this work, we mainly improve a U-shaped Encoder-Decoder model (we take it as the baseline model, which has the same architecture as the original U-Net, but contains smaller number of channels than the original U-Net) from three aspects: integrating residual blocks into the baseline model to deepen the model appropriately, so as to improve the feature extraction ability of the model; constructing hybrid attention blocks to improve the feature representation ability of the decoder part; using depthwise separable convolution instead of conventional convolution to reduce the parameters of the model for facilitating the deployment of the model to embedded devices. To validate these mechanisms, we conduct various ablation studies on CamCrack789 to demonstrate their effectiveness. The experimental results of different blocks are shown in Table IV, where RB represents the residual blocks, HAB is the hybrid attention block, and DS denotes the depthwise separable convolution. From Table IV, we can notice that both Baseline+RB and Baseline+HAB achieve better results than the Baseline model in terms of the comprehensive measurement indicator F1, indicating that both residual blocks and hybrid attention mechanisms help to improve the performance of the Baseline model. In addition, the proposed RHA-Net outperforms the Baseline model by 1.89\% and 1\% in terms of the Pr and F1, respectively, demonstrating that integrating both residual blocks and hybrid attention mechanisms into the Baseline model can further improve its performance.

\begin{table} 
\renewcommand\arraystretch{1.3}
    \centering
    \setlength{\tabcolsep}{2.8mm}{
    \caption{Results of ablation experiment on CamCrack789}
   \begin{tabular}{lcccc}
    \toprule

   Models & Pr & Re & F1 \\
    \midrule
   Baseline & 0.9256 &  0.9661 & 0.9430 \\
   Baseline+RB & 0.9236 & {\bfseries 0.9728} & 0.9458 \\
   Baseline+HAB & 0.9426 & 0.9620 & 0.9501  \\
			\midrule
   Baseline+RB+HAB (RHA-Net) & {\bfseries 0.9445} & 0.9654 & {\bfseries 0.9530} \\
   Baseline+RB+HAB+DS (RHA-Net$^{*}$) & 0.9416 & 0.9628 & 0.9501 \\
    \bottomrule
   \end{tabular}}
    \label{tab: ablation}
\end{table}

\begin{table}
  \renewcommand\arraystretch{1.3}
  \centering
  \setlength{\tabcolsep}{4.6mm}{
  \caption{Comparison of parameters and FLOPs of all methods}
	\begin{tabular}{lccc}
    \toprule
    Models & Params & FLOPs \\
    \midrule
			 FCN \cite{long2015fully} & 18.64 M  & 239.46 G  \\
    U-Net \cite{ronneberger2015u} & 17.25 M & 375.24 G  \\ 
    Attention U-Net \cite{oktay2018attention} & 34.88 M & 624.72 G \\
    DeepCrack \cite{zou2018deepcrack} & 30.91 M & 1283.64 G \\
    DeepCrack \cite{liu2019deepcrack} & 14.72 M & 188.56 G\\
 				\midrule
    RHA-Net  & {\bfseries 1.67 M} & {\bfseries 21.6 G} \\
				RHA-Net$^{*}$ & {\bfseries 0.57 M} & {\bfseries 9.68 G} \\
    \bottomrule
	\end{tabular}}
	\label{tab:speed}
\end{table}

\begin{figure*}[!ht]
    \centering
    \includegraphics[width=18.2cm]{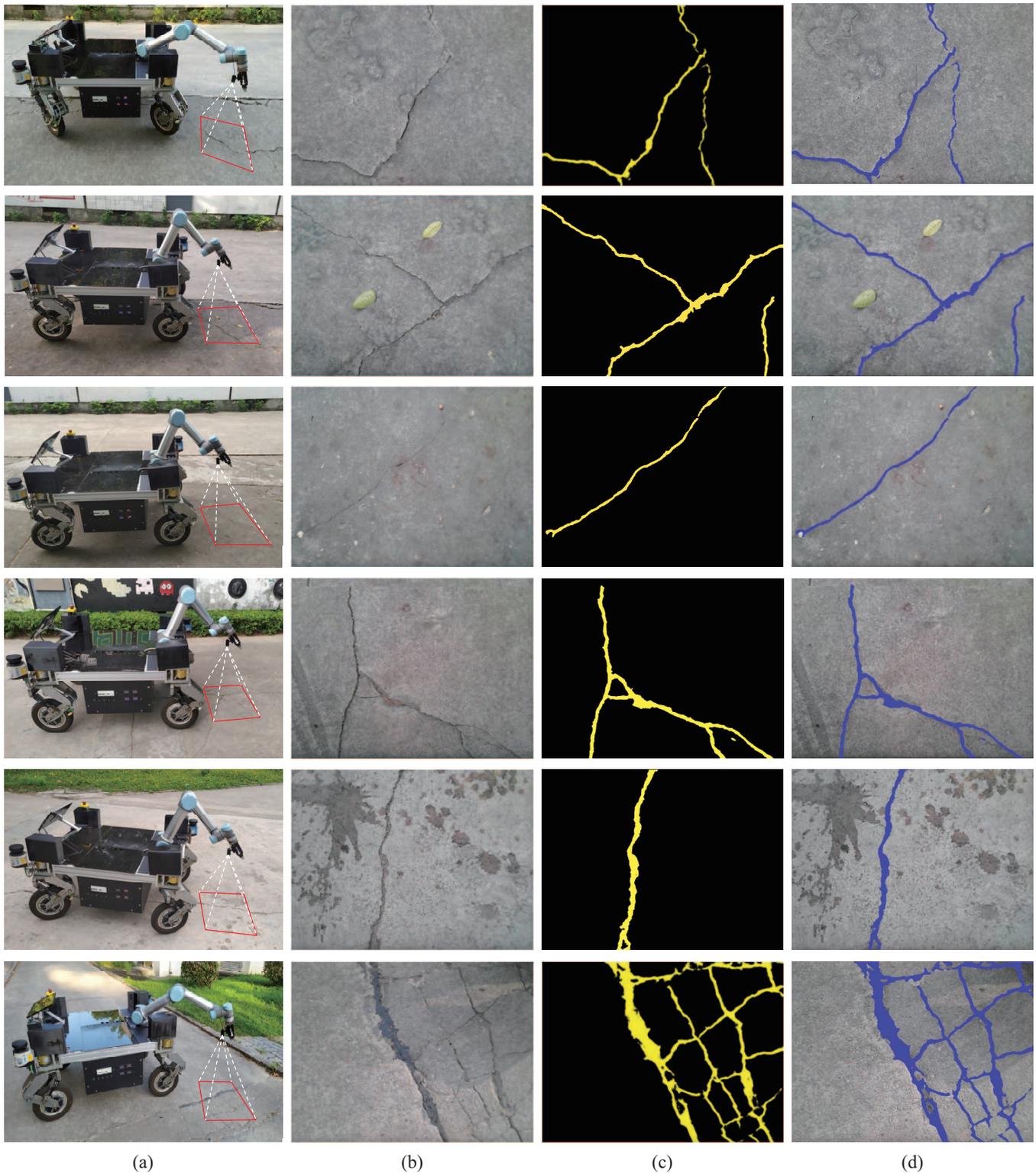}
    \caption{Examples of real-time segmentation results of pavement cracks in different road scenarios. (a) Road scenarios; (b) Raw images; (c) Segmentation results; (d) Overlaid images of the raw images and segmentation results.}
    \label{fig:real-time}
\end{figure*}

\subsection{Results of Real Time Deployment}
\label{sec:application}
To deploy the proposed model to embedded devices, we use depthwise separable convolution to replace conventional convolution of Baseline+RB+HAB except that of HAB. The processing time of the proposed model is compared with other models on all datasets, with the results provided in Table I to Table IV. It can be seen from Table I to Table IV that, the proposed RHA-Net is faster than other comparison methods and achieves the best performance. In addition, the light-weighted version RHA-Net* is not only the fastest (more than 2 times faster than U-Net) among all the models, but also obtains a commendable performance similar to that of RHA-Net with fewer network parameters on four datasets. Table VI illustrates the numbers of model parameters and floating points operations per second (FLOPs) of all models. The two metrics are computed based on an input size of 3$\times$640$\times$480. The parameters and FLOPs of the proposed RHA-Net is 1.67 MB and 21.6 GB, which is about 90.31\% and 94.24\% reduction compared to the 17.25 MB parameters and 375.24 GB FLOPs for U-Net respectively. In addition, the parameters and FLOPs of RHA-Net* is 0.57 MB and 9.68 GB, which is 65.86\% and 55.18\% reduction compared to the 1.67 MB parameters and 9.68 GB FLOPs for RHA-Net respectively. As RHA-Net* uses depthwise separable convolution instead of conventional convolution, which significantly decreases the model parameters and computational expense. In summary, the proposed model obtains a substantially fewer parameters and higher computational effciency compared with other state of the art crack segmentation models. Moreover, RHA-Net* gets a better balance between accuracy and computing complexity, so as to facilitate model deployment to embedded devices.

To illustrate the practicability of the proposed model, we deploy RHA-Net$^{*}$ trained on CamCrack789 to Jetson TX2 mounted on a mobile robot, and carry out a real test on campus road. As a result, the mobile robot system for pavement crack segmentation reaches about 25 FPS on Jetson TX2. Visualization results of pavement cracks segmented by a self-designed mobile robot on the campus road are shown in Fig.~\ref{fig:real-time}. From the first and second row of Fig.~\ref{fig:real-time}, we can see that RHA-Net* can not only segment the coarse cracks, but also perform well in segmenting the tiny cracks. The second to fifth row of Fig.~\ref{fig:real-time} illustrate crack images with a variety of features on the pavement, containing tree leaves, branches, rut marks, and water stains. It can be seen that RHA-Net* can accurately segment pavement cracks and cause very few false positives, which means that it can effectively localize pavement cracks and well suppress the influence of those background noises. 

Although the proposed model has improved the accuracy of pavement crack segmentation, it still faces challenges in segmenting cracks with very complex topology. As shown in the last row of Fig.~\ref{fig:real-time}, some cracks near the edge of the image are not well detected by the network, which may be caused by the motion of mobile robot and uneven illumination. In addition, we note that a sealed crack connected to other cracks is misidentified as a crack. This is because the training data does not include any crack images containing this features. Therefore, there is still room for improving segmentation performance for complex cracks.

\section{Conclusion}
\label{sec:Conclusion}

In this work, a pavement crack segmentation network RHA-Net based on the residual blocks and hybrid attention blocks is proposed. We use the residual blocks to enhance the extraction capability of the proposed model for crack features. The hybrid attention blocks are embedded between the encoder and decoder network to effectively integrate information from low-level feature maps and high-level ones, so as to improve the feature representation ability of the proposed method for tiny cracks and suppress background interference. Extensive experiments indicate that the proposed network can achieve better performances than the comparison methods. The proposed network obtains the best F1 value and the fastest inference speed on the four datasets. Ablation experiments illustrate the effectiveness of the residual blocks and hybrid attention blocks. In addition, the light-weighted version RHA-Net* is generated by replacing the conventional convolution with depthwise separable convolution. Results of pavement crack images acquired by a self-designed mobile robot in real scenarios further demonstrate the effectiveness and efficiency of RHA-Net*.

\bibliographystyle{ieeetran}
\bibliography{crack.bib}

\end{document}